\documentclass[letterpaper, 10pt, conference]{ieeeconf}
\IEEEoverridecommandlockouts  
\usepackage{algorithm}
\usepackage{amsmath}
\usepackage{amssymb}
\usepackage{bm}
\usepackage{color}
\usepackage{graphicx}
\usepackage[colorlinks, linkcolor=blue, urlcolor=blue]{hyperref}
\newcommand{\defeq}{:=}

\def\bfcd{\dot{\bfc}}

\def\bfxid{\dot{\bfxi}}

\def\omegad{\dot{\omega}}

\def\xid{\dot{\xi}}

\def\bfcdd{\ddot{\bfc}}

\newcommand{\bfsigma}{\boldsymbol{\sigma}}
\newcommand{\bftau}{\boldsymbol{\tau}}

\newcommand{\bfxi}{\boldsymbol{\xi}}

\newcommand{\bfb}{\ensuremath {\bm{b}}}
\newcommand{\bfc}{\ensuremath {\bm{c}}}
\newcommand{\bfd}{\ensuremath {\bm{d}}}
\newcommand{\bfe}{\ensuremath {\bm{e}}}
\newcommand{\bff}{\ensuremath {\bm{f}}}
\newcommand{\bfg}{\ensuremath {\bm{g}}}
\newcommand{\bfh}{\ensuremath {\bm{h}}}

\newcommand{\bfn}{\ensuremath {\bm{n}}}

\newcommand{\bfp}{\ensuremath {\bm{p}}}

\newcommand{\bfu}{\ensuremath {\bm{u}}}
\newcommand{\bfv}{\ensuremath {\bm{v}}}
\newcommand{\bfw}{\ensuremath {\bm{w}}}
\newcommand{\bfx}{\ensuremath {\bm{x}}}

\newcommand{\bfz}{\ensuremath {\bm{z}}}

\newcommand{\bfA}{\mathbf{A}}

\newcommand{\bfC}{\mathbf{C}}

\newcommand{\bfG}{\mathbf{G}}

\newcommand{\bfL}{\mathbf{L}}

\newcommand{\bfR}{\mathbf{R}}

\newcommand{\bfW}{\mathbf{W}}

\newcommand{\calX}{{\cal X}}

\def\bfRt{\bar{\bfR}}
\def\bfzt{\bar{\bfz}}
\def\eye{\boldsymbol{I}_3}
\def\fmax{f_\textit{max}}
\def\fmin{f_\textit{min}}
\def\hmax{h_\textit{max}}
\def\hmin{h_\textit{min}}
\def\kp{\ensuremath k_\textit{p}}
\def\lambdamax{\lambda_\textit{max}}
\def\lambdamin{\lambda_\textit{min}}

\def\xiznext{\xi_z^\textit{next}}
\def\zeros{\boldsymbol{0}}
\def\zt{\bar{z}}

\title{\LARGE \bf
    Biped Stabilization by Linear Feedback of the \\
    Variable-Height Inverted Pendulum Model
}

\author{St\'ephane Caron%
    \thanks{The author is with ANYbotics AG, Switzerland. This work was carried
    out while he was with the Montpellier Laboratory of Informatics, Robotics
    and Microelectronics (LIRMM), CNRS--University of Montpellier, France.
    E-mail: {\tt\footnotesize stephane.caron@normalesup.org}}%
}

\begin{document}

\maketitle
\thispagestyle{empty}
\pagestyle{empty}

\begin{abstract}
    The variable-height inverted pendulum (VHIP) model enables a new balancing strategy by height variations of the center of mass, in addition to the well-known ankle strategy. We propose a biped stabilizer based on linear feedback of the VHIP that is simple to implement, coincides with the state-of-the-art for small perturbations and is able to recover from larger perturbations thanks to this new strategy. This solution is based on ``best-effort'' pole placement of a 4D divergent component of motion for the VHIP under input feasibility and state viability constraints. We complement it with a suitable whole-body admittance control law and test the resulting stabilizer on the HRP-4 humanoid robot.
\end{abstract}

\section{Introduction}

Legged robots are constantly compensating undesired motions of their floating base by regulating their interaction forces with the environment, an action known as \emph{balancing} or \emph{stabilization}. Stabilization can be implemented by a collection of feedback control laws, referred to collectively as \emph{stabilizer}. Any stabilizer needs to answer two core questions. First, what contact wrench should be applied onto the environment to correct the floating base position and motion? Second, how to realize this contact wrench?

Reduced models play a key role in answering the first question. A reduced model parameterizes the contact wrench by at most six variables. The most common reduced model is the \emph{linear inverted pendulum}~(LIP)~\cite{kajita2001iros}, which assumes constant centroidal angular momentum as well as a planar motion of the center of mass~(CoM), and parameterizes the contact wrench by a two-dimensional zero-tilting moment point~(ZMP)~\cite{vukobratovic1972}. For this model, the answer to our question is known: the ZMP of ground reaction forces should react proportionally to deviations of the \emph{divergent component of motion}~(DCM) of the floating base. This solution maximizes the basin of attraction among linear feedback controllers~\cite{sugihara2009icra} and has been widely reproduced~\cite{kajita2010iros, englsberger2011iros, koolen2016ijhr, caron2019icra}.

The LIP leaves us with two avenues for improvement: enabling angular momentum~\cite{wiedebach2016humanoids} or height variations. The \emph{variable-height inverted pendulum}~(VHIP)~\cite{koolen2016humanoids} explores the latter with the addition of an input $\lambda$ that represents the stiffness of the massless leg between CoM and ZMP. This new input makes the system nonlinear, but gives it the ability to fall or push harder on the ground, enabling a new ``height variation'' recovery strategy when ZMP compensation is not enough~\cite{koolen2016humanoids}. Studies of the VHIP have focused on using this strategy to balance in the 2D sagittal plane~\cite{koolen2016humanoids, pratt2007icra, ramos2015humanoids}, with recent applications tested on hardware~\cite{vanhofslot2019ral}. Numerical optimization of 3D VHIP trajectories has also turned out to be tractable for both balancing and walking~\cite{feng2013humanoids, caron2019tro}, with extensions that add angular momentum variations~\cite{guan2019iros}.

The main alternative to the VHIP is the well-known 3D DCM~\cite{englsberger2015tro}. This reduced model works with the same set of assumptions, but is more tractable owing to its linear dynamics. Notably, it can be used for linear feedback control, whereas the aforementioned 3D VHIP controllers are based on nonlinear model predictive control. The price that the 3D DCM pays for this simplicity lies in its nonlinear feasibility constraints~\cite{caron2017iros}. Unless they are taken into account, \emph{e.g.} by nonlinear model predictive control, the 3D DCM does not yield the height-variation strategy that the VHIP allows.

\begin{figure*}[t]
    \centering
    \vspace{1em}
    \href{https://www.youtube.com/watch?v=vFCFKAunsYM&t=20}{\includegraphics[width=1.8\columnwidth]{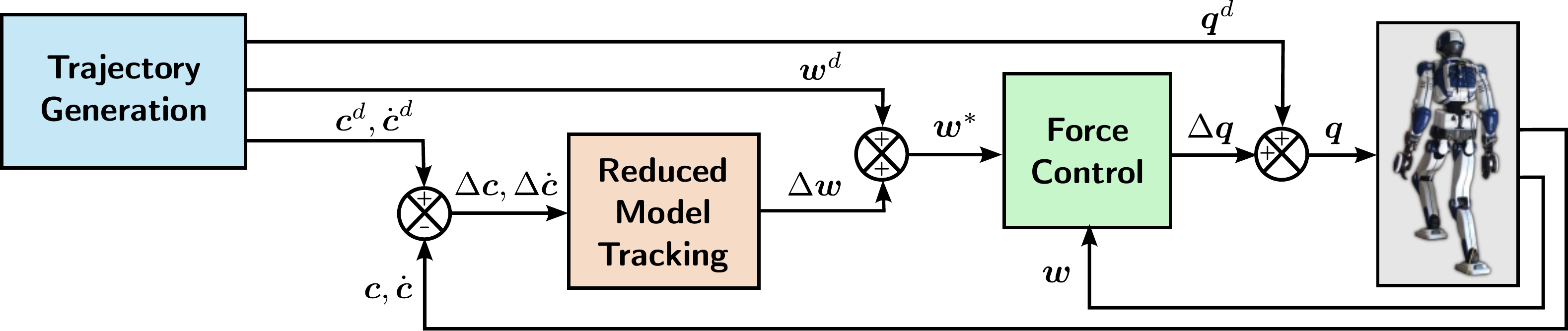}}
    \caption{
        Overview of the standard stabilization pipeline. A planning component, typically based on trajectory optimization, provides dynamically-consistent references $\square^d$ for all quantities. Reduced model tracking evaluates the errors $\Delta \bfc, \Delta \bfcd$ in center of mass tracking (previously combined into the capture point of the LIP, extended to a 4D DCM of the VHIP in Section~\ref{sec:vhip-tracking}) and outputs a corresponding modification $\Delta \bfw$ of the commanded contact wrench $\bfw^*$. Force control converts this wrench into joint angles (extended to include height variations in Section~\ref{sec:force}) for our position-controlled robot.
    }
    \label{fig:overview}
\end{figure*}

In this study, we uncover a new solution on the spectrum:~a linear feedback controller for the VHIP that coincides with the 3D DCM as long as feasibility constraints are not saturated, and naturally resorts to height variations when the ZMP approaches the edge of its support area. This component generalizes LIP tracking~\cite{kajita2010iros} to VHIP tracking in the standard stabilization pipeline (Figure~\ref{fig:overview}).

To implement VHIP tracking on our position-controlled robot, we also propose a whole-body admittance control strategy corresponding to the new input $\lambda$. We validate the closed-loop stability and performance of the resulting stabilizer in experiments with the HRP-4 humanoid.

\section{Reduced model tracking}

What reaction forces should the robot apply on its environment to correct the position of its floating base? 

Let us consider the net contact wrench $(\bff, \bftau_{\bfc})$, which consists of the resultant $\bff$ of contact forces applied to the robot and their moment $\bftau_{\bfc}$ around the center of mass (CoM) $\bfc$. The equations of motion of an articulated robot consist of two parts: joint dynamics, and floating base dynamics~\cite{wieber2016modeling}. The latter are governed by Newton and Euler equations:
\begin{equation}
    \label{eq:newton-euler}
    \begin{bmatrix} m \bfcdd \\ \dot{\bfL}_{\bfc} \end{bmatrix}
    =
    \begin{bmatrix} \bff \\ \bftau_{\bfc} \end{bmatrix}
    +
    \begin{bmatrix} m \bfg \\ \bm{0} \end{bmatrix}
\end{equation}
where $m$ denotes the total mass, $\bfg$ is the gravity vector, $\bfc$ the position of the center of mass (CoM) and $\bfL_{\bfc}$ the angular momentum around $\bfc$.  

\subsection{Inverted pendulum models}

\subsubsection{Variable-height inverted pendulum} assuming constant centroidal angular momentum $\dot{\bfL}_{\bfc} = \bm{0}$, centroidal dynamics~\cite{orin2013ar} are reduced to the variable-height inverted pendulum~(VHIP) model~\cite{koolen2016humanoids}:
\begin{equation}
    \label{eq:vhip}
    \bfcdd = \lambda (\bfc - \bfz) + \bfg
\end{equation}
With its angular coordinates constrained to the manifold $\bftau_{\bfc} = \bm{0}$, the contact wrench is characterized by the three coordinates $(\lambda, \bfz)$ that define its resultant $\bff = m \lambda (\bfc - \bfz)$. The coordinate $\lambda$ is a normalized stiffness while $\bfz$ is the zero-tilting moment point (ZMP)~\cite{vukobratovic1972}. To be feasible, a contact wrench must have positive $\lambda$ (contact unilaterality) and a ZMP within the contact surface; mathematically, these are linear constraints $\lambda > 0$ and $\bfC \bfz \leq \bfd$. Note that, although we write it here as a 3D vector in the world frame, the ZMP is a two-dimensional quantity as it lies on the contact surface.

\subsubsection{Linear inverted pendulum} when walking over a horizontal surface, further assuming a constant CoM height $c_z = h$ leads to the linear inverted pendulum~(LIP) model~\cite{kajita2001iros}:
\begin{equation}
    \label{eq:lip}
    \bfcdd = \omega_0^2 (\bfc - \bfz) + \bfg
\end{equation}
where $\omega_0 = \sqrt{g / h}$ is the natural frequency of the LIP, with $g$ the standard acceleration due to gravity. With its stiffness $\lambda$ constrained to the manifold $\lambda = \omega_0^2$, the contact wrench is then characterized by the two coordinates $\bfz$ of the ZMP.

\subsubsection{Floating-base inverted pendulum} alternatively, the CoM acceleration can be parameterized by the \emph{enhanced centroidal moment pivot}~(eCMP)~\cite{englsberger2015tro}:
\begin{equation}
    \label{eq:fip}
    \bfcdd = \frac{1}{b^2} (\bfc - \bfe) + \bfg
\end{equation}
where $b > 0$ is a new parameter chosen by the user. In the LIP where $b = 1/\omega_0$, the eCMP $\bfe$ coincides with the ZMP $\bfz$. While the parameter $b$ is usually chosen close to $1 / \omega_0$, the interest of this model is that it does not include a CoM height constraint. The CoM is free to move vertically, but then the eCMP leaves the contact area. The ZMP is then located at the intersection of the ray from CoM to eCMP with the contact area.

The floating-base inverted pendulum (FIP) expresses the same contact wrenches as the VHIP but its input is the eCMP $\bfe$ rather than the pair $(\lambda, \bfz)$. The price to pay for this simpler input is its ZMP feasibility constraint $\bfC \bfz \leq \bfd$ becomes nonlinear in $\bfe$~\cite{caron2017iros} (it can be approximated linearly for small height variations~\cite{zamparelli2018ifac}).

\subsection{Divergent components of motion}

For all three reduced models, an exponential dichotomy can be applied to decompose the second-order dynamics of the center of mass into two first-order systems.

\subsubsection{Linear inverted pendulum} we can define the {divergent component of motion}~(DCM), also known as {capture point}~\cite{pratt2006humanoids} for the LIP, as:
\begin{align}
    \label{eq:lip-dcm}
    \bfxi & = \bfc + \frac{\bfcd}{\omega_0} 
\end{align}
Taking the time derivative of this expression and injecting Equation~\eqref{eq:lip} yields:
\begin{align}
    \bfxid & = \omega_0 (\bfxi - \bfz) + \frac{\bfg}{\omega_0} &
    \bfcd & = \omega_0 (\bfxi - \bfc)
    \label{eq:lip-decoupling}
\end{align}
The DCM is repelled by the ZMP, while the CoM is attracted to the DCM. Notably, CoM dynamics have become both independent from the ZMP and stable with respect to the DCM. Controlling only the DCM therefore suffices to control the CoM, and thus the floating base of the robot.

\subsubsection{Floating-base inverted pendulum} the 3D DCM is defined for the FIP model as:
\begin{align}
    \label{eq:fip-dcm}
    \bfxi & = \bfc + b \bfcd
\end{align}
Using Equation~\eqref{eq:fip}, the time derivative of this expression is:
\begin{align}
    \bfxid & = \frac{1}{b} (\bfxi - \bfe) + b \bfg&
    \bfcd & = \frac{1}{b} (\bfxi - \bfc)
    \label{eq:fip-decoupling}
\end{align}
Second order dynamics are again decoupled in two linear first-order DCM--eCMP and CoM--DCM sub-systems.

\subsubsection{Variable height inverted pendulum} a divergent component of motion for the VHIP can be defined as~\cite{hopkins2014humanoids}:
\begin{align}
    \label{eq:vhip-dcm}
    \bfxi & = \bfc + \frac{\bfcd}{\omega}
\end{align}
where the natural frequency $\omega$ is now \emph{time-varying} and satisfies the Riccati equation:
\begin{align}
    \label{eq:riccati}
    \omegad & = \omega^2 - \lambda
\end{align}
See \emph{e.g.}~\cite{caron2019tro} for details on how this equation appears in the derivation of this DCM, or~\cite{garcia2019arxiv, garcia-chavez2019humanoids} for a similar approach based on the natural period $T = \omega^{-1}$. Taking the time derivative of~\eqref{eq:vhip-dcm} and injecting Equations~\eqref{eq:vhip} and \eqref{eq:riccati} yields:
\begin{align}
    \bfxid & = \frac{\lambda}{\omega} (\bfxi - \bfz) + \frac{\bfg}{\omega} &
    \bfcd & = \omega (\bfxi - \bfc)
    \label{eq:vhip-decoupling}
\end{align}
Second-order dynamics are thus decoupled in two first-order \emph{nonlinear} systems.

\subsection{Linear feedback control of the DCM}
\label{sec:pole-placement}

Let us denote with the superscript $\square^d$ a reference state of the reduced model satisfying all of its equations, for instance $\bfcd^d = \omega^d (\bfxi^d - \bfc^d)$. This reference can be obtained in a prior step by trajectory generation~\cite{englsberger2017iros} or optimization~\cite{feng2013humanoids, caron2019tro, winkler2017ral, mastalli2019arxiv} (Figure~\ref{fig:overview}). The error on a quantity $\bfx$ is written $\Delta \bfx \defeq \bfx - \bfx^d$. In the case of the LIP, from Equation~\eqref{eq:lip-decoupling} the time derivative of the DCM error $\Delta \bfxi$ is:
\begin{align}
    \label{eq:deriv1}
    \Delta \bfxid & = \omega_0 (\Delta \bfxi - \Delta \bfz)
\end{align}
We want to realize \emph{pole placement}~\cite{morisawa2012humanoids} so that this error converges exponentially to zero:
\begin{align}
    \Delta \bfxid & = (1 - \kp) \omega_0 \Delta \bfxi
\end{align}
where $1 - \kp < 0$ is the normalized closed-loop pole. Combining these two equations yields:
\begin{align}
    \label{eq:deriv3}
    \Delta \bfz & = \kp \Delta \bfxi
\end{align}
This control law answers our initial question and implements LIP tracking. When it observes a deviation $\Delta \bfxi$ of its DCM from the reference $\bfxi^d$, the robot modifies (the ZMP of) its contact wrench in proportion to the DCM error. An integral term can also be added to eliminate steady-state error~\cite{morisawa2012humanoids}.

The same derivation applied to the FIP~\cite{englsberger2015tro} yields:
\begin{align}
    \Delta \bfxid & = (1 - \kp) \frac{1}{b} \Delta \bfxi & \Delta \bfe & = \kp \Delta \bfxi 
\end{align}
Pole placement is thus generalized to the 3D DCM parameterized by $b$. As long as the eCMP $\bfe = \bfe^d + \Delta \bfe$ is feasible, \emph{i.e.} the ZMP projected along the CoM--eCMP axis lies in the support area, it achieves optimal closed-loop dynamics. When the corresponding ZMP falls outside of the support area, it is projected back to it and closed-loop pole placement is not guaranteed any more. In this case, the DCM error will either decrease sufficiently to end saturation, or diverge.

\section{Linear feedback control of the VHIP}
\label{sec:vhip-tracking}

Recent studies of the VHIP showed the existence of an alternative: even with the ZMP constrained at the boundary of its support area, the system might be balanced using the height-variation strategy. Trajectories that display this strategy can be found by numerical optimization~\cite{feng2013humanoids, ramos2015humanoids, koolen2016ijhr, caron2019tro}, but not by proportional feedback of the 3D DCM~\eqref{eq:fip-dcm} as they correspond to variations of the parameter $b$. In what follows, we will see this behavior emerge from linear feedback control of a different DCM of the VHIP. 

\subsection{Four-dimensional DCM for the VHIP}

We noted previously~\cite{caron2019tro} how $\omega$ behaves like a divergent component repelled by the input $\lambda$. Let us embrace this observation fully and consider the joint vector $\bfx = [\bfxi\,; \omega]$ as a four-dimensional DCM with three spatial and one frequential coordinate. Its time derivatives are given by Equations \eqref{eq:riccati}--\eqref{eq:vhip-decoupling}. Let us take their first order differentials with respect to the reference. For the natural frequency, we get:
\begin{align}
    \Delta \omegad
    & = (\omega^d + \omega) \Delta \omega - \Delta \lambda
    \approx 2 \omega^d \Delta \omega - \Delta \lambda
    \label{eq:delta-omega}
\end{align}
where we assume that quadratic and higher order errors such as $\Delta \omega^2$ can be neglected. Applying a similar derivation to the spatial DCM $\bfxi$ yields:
\begin{align}
    \Delta \bfxid & = \frac{\lambda^d}{\omega^d} (\Delta \bfxi - \Delta \bfz) + \frac{\bfxi^d - \bfz^d}{\omega^d} \Delta \lambda - \frac{\bfcdd^d}{{\omega^d}^2} \Delta \omega \label{eq:delta-xi}
\end{align}
Bringing these two equations together, we obtain a linearized state-space model
for the four-dimensional state $\Delta \bfx = [\Delta \bfxi\,; \Delta \omega]$ with three-dimensional input $\Delta \bfu = [\Delta \bfz\,; \Delta \lambda]$. We assign the closed-loop poles of this system as:
\begin{align}
    \Delta \bfxid & = (1 - \kp) \frac{\lambda^d}{\omega^d} \Delta \bfxi \label{eq:pole-xi} \\
    \Delta \omegad & = (1 - \kp) \omega^d \Delta \omega \label{eq:pole-omega}
\end{align}
Combining Equations~\eqref{eq:delta-omega}--\eqref{eq:delta-xi} and \eqref{eq:pole-xi}--\eqref{eq:pole-omega} yields:
\begin{align}
    \Delta \bfz + \frac{\bfxi^d - \bfv^d}{\omega^d} \Delta \omega + \frac{\bfz^d - \bfxi^d}{\lambda^d} \Delta \lambda & = \kp \Delta \bfxi \label{eq:delta-z} \\
    \Delta \lambda = \omega^d (1 + \kp) \Delta \omega \label{eq:delta-lambda}
\end{align}
where we used the shorthand $\bfv^d \defeq \bfz^d - \bfg^d / \lambda^d$, a time-variant variant of the \emph{virtual repellent point} formula~\cite{englsberger2015tro}. This equation is central to our feedback controller: it shows how the inputs $\Delta \bfz$, $\Delta \lambda$ should be chosen as functions of the current state $\Delta \bfxi$, $\Delta \omega$ and feedback gain $k_p$. When $\Delta \omega = 0$~Hz and $\Delta \lambda = 0$~Hz$^2$, Equation~\eqref{eq:delta-z} simplifies into the standard proportional feedback of the DCM at the ZMP~\eqref{eq:deriv3}. Meanwhile, Equation~\eqref{eq:delta-lambda} provides a direct analogous of Equation~\eqref{eq:deriv3} over frequential coordinates. 

\subsection{The DCM is not a direct measurement}

A novelty of the VHIP lies in its ability to \emph{choose} its DCM by varying the natural frequency $\omega$. In the LIP or FIP, the DCM error $\Delta \bfxi$ is fully determined from sensory measurements $\Delta \bfc$ and $\Delta \bfcd$ by Equations~\eqref{eq:lip-dcm} and~\eqref{eq:fip-dcm}. In the VHIP, differentiating Equations~\eqref{eq:vhip-dcm} gives us:
\begin{align}
    \label{eq:vhip-dcm-omega}
    \Delta \bfxi & = \Delta \bfc + \frac{\Delta \bfcd}{\omega^d} - \frac{\bfcd}{\omega^d} \frac{\Delta \omega}{\omega^d}
\end{align}
The measured output vector $\Delta \bfc + \Delta \bfcd / \omega^d$ has dimension three, but the state vector $[\Delta \bfxi\,; \Delta \omega]$ has dimension four. The extra dimension is not an exogenous output: rather, the controller has an internal state by which it \emph{decides how to weigh} sensory measurements. Intuitively, if the robot is pushed hard enough so that the ZMP saturates its support area for the current value of $\omega$, the controller can choose to increase $\omega$ instead, thus increasing $\lambda$ and pushing harder on the ground by~\eqref{eq:delta-lambda}. This way, it can keep the spatial DCM in the vicinity of the contact area, yet only for a while as pushing harder on the ground requires raising the CoM, which is only available in limited supply depending on joint kinematic and torque limits.

\subsection{Input feasibility conditions}

To generate feasible contact wrenches, the inputs $\Delta \bfz$ and $\Delta \lambda$ need to satisfy a set of inequality constraints.

Let us define the ZMP frame as the average contact frame over all contacts with the environment. We denote the origin of this frame by $\bfp$ frame and its rotation matrix (from ZMP to inertial frame) by $\bfR$.

\subsubsection{ZMP support area} the coordinates of the ZMP compensation in the inertial frame are then given by $\Delta \bfz = \bfRt \bfzt$, where $\bfzt \in \mathbb{R}^2$ and $\bfRt$ consists of the first two columns of $\bfR$. In single support, the ZMP after compensation should lie within the contact area, so that:
\begin{align}
    -X \leq \zt^d_x + \Delta \zt_x \leq X \\
    -Y \leq \zt^d_y + \Delta \zt_y \leq Y
\end{align}
The inequalities provide a halfspace representation:
\begin{align}
    \label{ineq:zmp}
    \bfC \Delta \bfzt \leq \bfd
\end{align}
In double support, and more generally in multi-contact scenarios, similar halfspace representations of the multi-contact ZMP support area can be obtained by projection of the contact wrench cone~\cite{caron2016tro}. A simple method to compute it is reported in Section IV.C of~\cite{caron2016humanoids}.

\subsubsection{Actuation limits} contact unilaterality and joint torque limits of the underlying robot model can be approximated in the reduced model by lower and upper bounds on the normal contact force:
\begin{align}
    \fmin \leq (\bfn \cdot \bff) = m \lambda \bfn \cdot (\bfc - \bfp) \leq \fmax
\end{align}
These inequalities can be readily rewritten:
\begin{align}
    \label{ineq:lambda}
    \frac{\fmin}{m \bfn \cdot (\bfc - \bfp)}
    \leq \lambda^d + \Delta \lambda
    \leq \frac{\fmax}{m \bfn \cdot (\bfc - \bfp)}
\end{align}
The lower bound $\lambdamin$ and upper bound $\lambdamax$ thus defined depend on actuation limits, total mass and the instantaneous position of the center of mass.

\subsection{State viability conditions}

Input feasibility conditions are not sufficient to guarantee that the system will not diverge to a failed state: they should be complemented by state viability conditions. Instances of viability conditions include keeping the capture point (for the LIP) inside the convex hull of ground contact points~\cite{sugihara2009icra}, or bounding joint accelerations in whole-body control to maintain joint angle limits in the long run~\cite{delprete2018ral}.

\subsubsection{Frequency limits} to be feasible, the natural frequency $\omega$ of the VHIP should not exceed the bounds of its corresponding input $\lambda$ (again in fitting analogy to the spatial DCM and ZMP support area):
\begin{align}
    \label{ineq:omega}
    \sqrt{\lambdamin} \leq \omega^d + \Delta \omega \leq \sqrt{\lambdamax}
\end{align}
The intuition for this viability condition lies in the Riccati equation~\eqref{eq:riccati}: once $\omega^2$ decreases below $\lambdamin$ it is impossible for $\omegad$ to be positive again. See Property~6 in~\cite{caron2019tro} for details.

\subsubsection{DCM height limits} variations of $\lambda$ require the underlying robot model to adjust the height of its CoM. From Equation~\eqref{eq:vhip-decoupling}, the CoM is attracted to the DCM, so that bounding DCM height is a sufficient condition to bound CoM height. Let us define:
\begin{align}
    \label{eq:xi-z-next}
    \xiznext & = \xi_z^d + \Delta \xi_z + (1 + \kappa) {\rm d}t \Delta \xid_z
\end{align}
where ${\rm d}t$ is the control period and $\kappa = 0.5$ is a damping factor to allow sliding on the constraint when it is saturated without making the control problem infeasible. Height limits are finally expressed as:
\begin{align}
    \hmin \leq \xiznext \leq \hmax
\end{align}

\subsection{Quadratic programming formulation}

Our problem is now specified: realize at best the desired closed-loop poles~\eqref{eq:pole-xi}--\eqref{eq:pole-omega} while satisfying input feasibility and state viability constraints. Let us cast it in matrix form\footnote{Before computing these matrices explicitly, we used (and warmly recommend) the CVXPY~\cite{cvxpy} modeling language to prototype our controller.} as a quadratic program:
\begin{align}
    \label{eq:qp-start}
    \underset{\calX}{\text{minimize}} \quad & \calX^T \bfW \calX \\
    \text{subject to} \quad
    & \bfG \calX \leq \bfh, \ \bfA \calX = \bfb
    \label{eq:qp-end}
\end{align}

We choose to include both states and inputs in our vector of optimization variables:
\begin{align}
    \calX = [\Delta \bfxi\ \Delta \omega\ \Delta \bfzt\ \Delta \lambda\ \Delta \bfsigma] \in \mathbb{R}^{10}
\end{align}
where $\Delta \bfsigma \in \mathbb{R}^3$ is an additional vector to allow violations of our desired pole placement on the spatial DCM. We make this vector homogeneous to a position:
\begin{align}
    \label{eq:add-sigma}
    \Delta \bfxid & = \frac{\lambda^d}{\omega^d} \left[ (1 - \kp) \Delta \bfxi + \Delta \bfsigma \right]
\end{align}

\subsubsection{Objective function} we minimize pole placement violations on horizontal components with highest priority, then on the vertical component:
\begin{align}
    \bfW & = \mathrm{diag}(
            \varepsilon,
            \varepsilon,
            \varepsilon,
            \varepsilon,
            \varepsilon,
            \varepsilon,
            \varepsilon,
            1,
            1,
            10^{-3})
\end{align}
where $\varepsilon \ll 10^{-3}$ makes the matrix $\bfW$ positive-definite, adding the minimization of other optimization variables as the lowest priority objective.

\subsubsection{Equality constraints} states and inputs are bound together by Equations~\eqref{eq:delta-z} (updated by~\eqref{eq:add-sigma} to include $\Delta \bfsigma$), \eqref{eq:delta-lambda} and \eqref{eq:vhip-dcm-omega}. Let us omit $\square^d$ superscripts for concision:
\begin{align}
    \bfA & = 
    \begin{bmatrix}
        -\kp \eye & \frac{\bfxi - \bfv}{\omega} & \bfRt & \frac{\bfz - \bfxi}{\lambda} & \eye \\
        \eye & \frac{\bfcd}{\omega^2} & \zeros_{3 \times 2} & \zeros_{3 \times 1} & \zeros_{3 \times 3} \\
        \zeros_{1 \times 3} & \omega (1 + \kp) & 0 & -1 & \zeros_{1 \times 3} \\
    \end{bmatrix} \\
    \bfb & = 
    \begin{bmatrix}
        \zeros_{3 \times 1} \ ;\ 
        \Delta \bfc + \frac{\Delta \bfcd}{\omega} \ ;\
        0
    \end{bmatrix}
\end{align}
where $\eye$ is the $3 \times 3$ identity matrix, $\bfA$ is a $7 \times 10$ matrix and $\bfb$ a $7 \times 1$ column vector.

\subsubsection{Inequality constraints} limits \eqref{ineq:zmp}~on the ZMP, \eqref{ineq:lambda}~on $\lambda$ and \eqref{ineq:omega}~on $\omega$ can be readily included as block matrices in $\bfG$ and $\bfh$. DCM height limits are obtained by injecting the expression of the related spatial DCM velocity~\eqref{eq:add-sigma} into Equation~\eqref{eq:xi-z-next}:
\begin{align}
    \bfG_{\xi}^{\textit{next}} & = 
    \begin{bmatrix}
        0 & 0 & +g_\xi & 0 & 0 & 0 & 0 & 0 & 0 & +g_\sigma \\
        0 & 0 & -g_\xi & 0 & 0 & 0 & 0 & 0 & 0 & -g_\sigma
    \end{bmatrix} \\
    \bfh_{\xi}^{\textit{next}} & =
    \begin{bmatrix}
        \hmax - \xi_z^d \ ;\  
        \xi_z^d - \hmin
    \end{bmatrix}
\end{align}
where $g_\sigma = (1 + \kappa) {\rm d}t {\lambda^d}/{\omega^d}$ and $g_\xi = 1 + g_\sigma (1 - \kp)$.

Overall, this quadratic program has 10 variables, 7 dense equality constraints and 6 + $m$ sparse inequality constraints, where the number $m$ of ZMP inequalities is usually less than $10$. During standing experiments with $m = 4$, it was solved in $0.1 \pm 0.05$~ms by LSSOL on a laptop computer. 

\subsection{Comparison to DCM--eCMP feedback control}

We compare the response of the best-effort pole placement QP~\eqref{eq:qp-start}--\eqref{eq:qp-end} with standard DCM--eCMP feedback control~\cite{koolen2016ijhr, englsberger2015tro} in a perfect simulation.\footnote{Source code: \url{https://github.com/stephane-caron/pymanoid/blob/master/examples/vhip\_stabilization.py}} The target state of the inverted pendulum is a static equilibrium with the center of mass $m = 38~\text{kg}$ located $h = 80$~cm above ground and 3~cm away from a lateral edge of the support area. Both controllers use the same feedback gain $\kp = 3$. The velocity scaling parameter of the DCM--eCMP feedback controller is set to the recommended value $b = \sqrt{h / g}$.

\begin{figure}[t]
    \includegraphics[width=0.99\columnwidth]{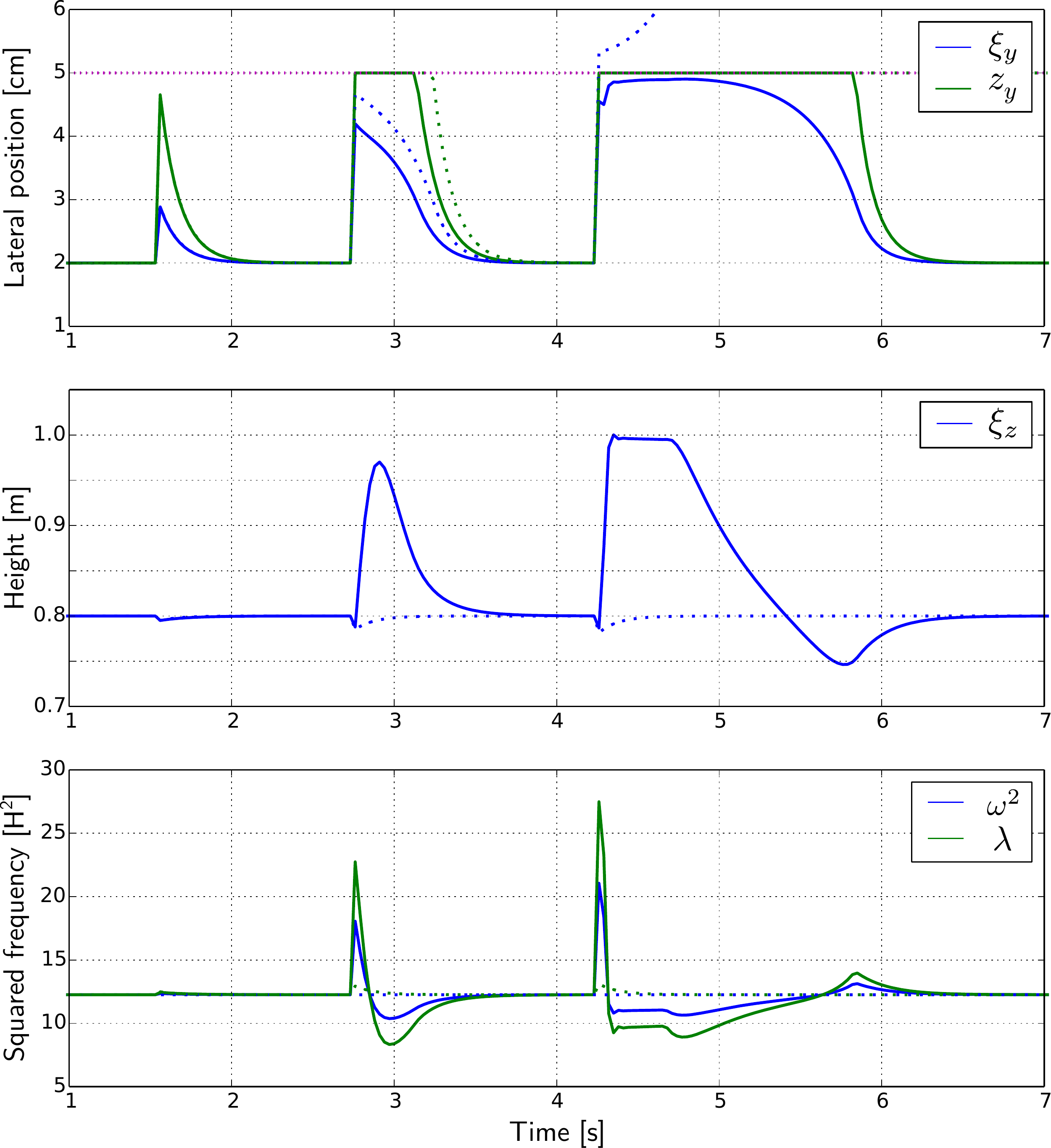}
    \caption{
        Comparison of the VHIP (solid lines) and DCM--eCMP (dotted lines) feedback controllers in perfect simulation. The DCM is impacted by impulses of increasing magnitude ($1.5~\text{N.s}$, $4.5~\text{N.s}$, $5.7~\text{N.s}$). For a small impulse, the two controllers match exactly. The VHIP controller is able to sustain larger impulses thanks to the height variation strategy.
    }
    \label{fig:simu}
\end{figure}

Figure~\ref{fig:simu} illustrates the response of the two controllers to increasingly high impulses applied to the CoM in the lateral direction. For a small impulse $i = m \Delta \dot{c}_y = 1.5~\text{N.s}$, the ZMP does not hit the edge of its support area and the two controllers match exactly.

For a medium impulse $i = 4.5~\text{N.s}$, the ZMP hits the edge of the area but the DCM is still inside it. The DCM--eCMP controller keeps on the edge until its DCM comes back in the vicinity of the desired state. The VHIP controller saturates its ZMP as well, and performs two additional behaviors:
\begin{itemize}
    \item \textbf{At impact time,} $\omega$ jumps from its reference $\omega_0 = 3.5~\text{Hz}$ to $4.2~\text{Hz}$. As a consequence, the post-impact DCM of the VHIP lies more inside the support area than its FIP counterpart.
    \item \textbf{After impact,} the controller adds around $15~\text{cm}$ of DCM height variations. As a consequence, the DCM is brought back to the desired state faster than its FIP counterpart.
\end{itemize}
Note how these two behaviors were \emph{not} explicitly part of our controller specification: they emerge from best-effort pole placement, input feasibility and state viability constraints.

For a larger impulse $i = 5.7~\text{N.s}$, the DCM of the FIP model falls outside of the support area. The DCM--eCMP controller is unable to recover from such disturbances. Applying the above strategy, the VHIP controller maintains its post-impact DCM within the support area. It then raises the DCM until the kinematic constraint $\xi_z \leq \hmax = 1~\text{m}$ is met. At this stage, state and inputs are fully saturated and the controller holds on. The DCM eventually comes back to the support area and returns to the desired state. In this particular example, failure thresholds are respectively $i = 5.2~\text{N.s}$ for the DCM--eCMP controller and $i = 6.0~\text{N.s}$ for the VHIP controller.

\section{Vertical force control}
\label{sec:force}

Reduced model control produces a desired net contact wrench. For torque-controlled robots, this net wrench is supplied as a target to whole-body control~\cite{ koolen2016ijhr, wensing2013icra, englsberger2018icra, bellicoso2018ral, dicarlo2018iros, kim2019highly}, and the resulting joint torques are sent to lower-level joint controllers. For position-controlled robots, an additional layer is required to regulate wrenches by admittance control.

\subsection{Whole-body admittance control}

Biped stabilizers usually include several admittance control laws in parallel, which can be collectively thought of as whole-body admittance control. There are two main approaches to regulate the net contact wrench: distribute it across end effectors in contact and regulate contact wrenches independently at each effector, or apply extra accelerations to the center of mass. These two approaches are not mutually exclusive. In what follows, we will use of the following:
\begin{itemize}
    \item \textbf{Foot damping control}~\cite{kajita2010iros} regulates the center of pressure under each foot by independent damping control over their respective ankle roll and pitch joints.
    \item \textbf{Foot force difference control}~\cite{kajita2010iros} regulates the \emph{difference} $(f_z^{\textit{left}} - f_z^{\textit{right}})$ of measured normal forces at each foot. Regulating a relative value elegantly avoids pitfalls coming from absolute discrepancies between measured and model forces (\emph{e.g.} in its model HRP-4 weighs 38~kg but our robot is now closer to 42~kg).
    \item \textbf{Horizontal CoM admittance control}~\cite{nagasaka1999thesis} improves net ZMP tracking by adding a horizontal CoM acceleration offset $\Delta \bfcdd_{\textit{xy}} = A_\textit{xy} \Delta \bfz$ proportional to the ZMP error.
\end{itemize}
See~\cite{caron2019icra} for a more detailed survey of the state of the art.

\subsection{Vertical CoM admittance control}

All admitance control strategies mentioned above contribute to improve ZMP tracking, which is consistent with the state of the art where reduced model control outputs a net ZMP $\bfz = \bfz^d + \Delta \bfz$. Using them altogether, the biped becomes compliant to external perturbations in the horizontal plane, while remaining totally stiff in the vertical direction. This phenomenon is illustrated with the LIP-based stabilizer from~\cite{caron2019icra} in the accompanying video.

To extend tracking to the VHIP, we want to track not only $\bfz$ but also $\lambda$, which corresponds to the normal contact force:
\begin{align}
    \label{eq:lambda-m}
    \lambda & = \frac{\bfn \cdot \bff}{m \bfn \cdot (\bfc - \bfp)} = \frac{f_z}{m (c_z - z_z)}
\end{align}

It seems we then need to regulate absolute forces, which in turns requires accurate force sensor calibration. This requirement was avoided by previous laws such as foot force difference control. We propose to circumvent it again by extending CoM admittance control to the vertical direction based on feedback, not from the net vertical force $f_z$, but from the normalized stiffness $\lambda$ of the VHIP:
\begin{align}
    \label{eq:vcac}
    \Delta \ddot{c}_z & = A_z \left(\lambda^d - \lambda\right) = -A_z \Delta \lambda
\end{align}
where $\lambda$ is measured from sensory data using Equation~\eqref{eq:lambda-m}.
\begin{figure}
    \includegraphics[width=0.99\columnwidth]{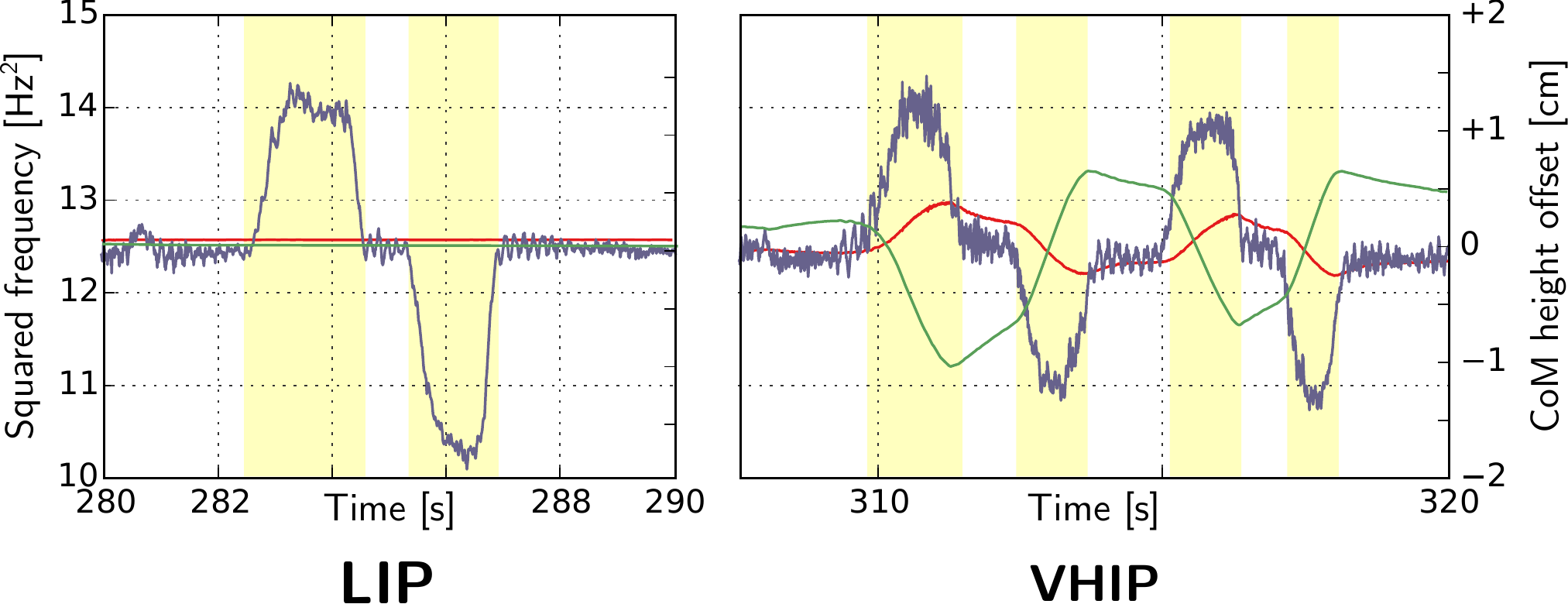}
    \caption{
        Response to vertical pushes for the LIP (left) and VHIP (right) stabilizers. User pushes (yellow background) are reflected in the measured normalized stiffness $\lambda$ (blue). In the LIP, the robot stays totally stiff. In the VHIP, its commanded stiffness $\lambda^c$ (red) increases and the CoM height (green) complies in the direction of the user's push.
    }
    \label{fig:admittance}
\end{figure}
This law is also consistent with the VHIP inputs. 

\section{Experimental validation}

We updated the stabilizer from~\cite{caron2019icra} with the two contributions of this manuscript: generalizing LIP tracking to VHIP tracking (Section~\ref{sec:vhip-tracking}) and extending CoM admittance control to the vertical direction (Section~\ref{sec:force}). We release the resulting controller as open source software\footnote{\url{https://github.com/stephane-caron/vhip\_walking\_controller/}} and invite readers to open issues in the corresponding repository for any related discussion. Closed issues also provide technical details not discussed in this manuscript.

\subsection{CoM admittance control}

We confirmed the stability of the closed-loop system consisting of both VHIP tracking and three-dimensional CoM admittance control by assessing the robot's compliance to external pushes, as shown in the accompanying video. In the horizontal plane, LIP and VHIP stabilizers perform identically, using the same gain $\kp = 1.4$ for both. In the vertical direction, the robot is totally stiff with the LIP and complies with the VHIP. Figure~\ref{fig:admittance} shows measured and commanded values of $\lambda$ from these experiments, as well as CoM height variations resulting from admittance control. When the robot is pushed down, $\lambda$ increases above $\lambda^c$ and the CoM complies downward.

\begin{figure}
    \centering
    \includegraphics[width=0.95\columnwidth]{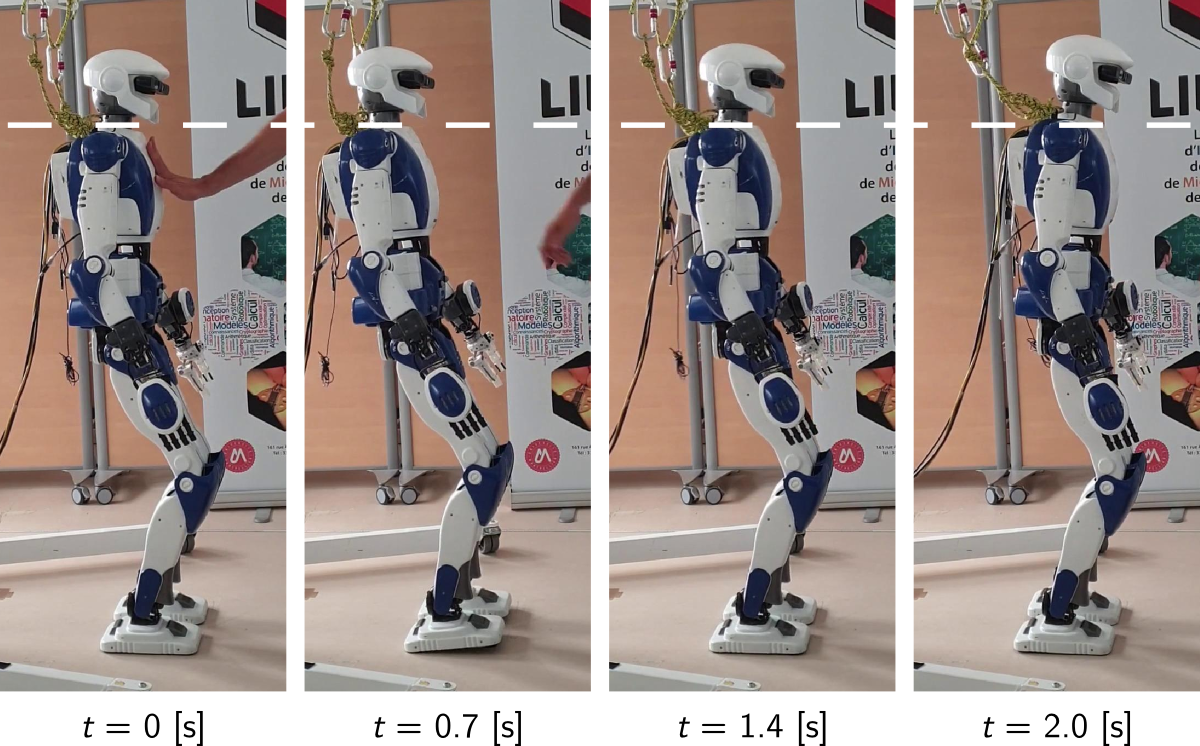}
    \caption{
        Push recovery with saturation of ZMP constraints shown in the accompanying video. The VHIP stabilizer behaves like its LIP counterpart while its commanded ZMP stays inside the support area, and resorts to height variations when the ZMP compensation strategy is exhausted. This hybrid behavior is not specified explicitly, but emerges from best-effort pole placement under feasibility and viability conditions.
    }
    \label{fig:experiment}
\end{figure}

We raised the admittance gain to $A_z = 0.005~\text{m}.\text{s}^{-2}$, achieving clear vertical compliance but with a rather low bandwidth. Increasing bandwidth using larger values of $A_z$ made the CoM prone to pick up vibrations from \emph{e.g.} force sensor noise, which are presently unfiltered. This behavior can be traded off with delay by applying signal filtering, \emph{e.g.} using the solution proposed in~\cite{englsberger2018icra}.

\subsection{Push recovery and walking}

The VHIP stabilizer on HRP-4 behaves essentially like its LIP counterpart until the ZMP hits the edge of the support area. In the experiment depicted in Figure~\ref{fig:experiment}, we trigger this event by pushing the robot until it rocks backward. The stabilizer then raises the CoM twice: a first time around $t=0.7~\text{s}$ to increase recovery forward acceleration, and a second time after $t=1.4~\text{s}$ when the robot is back on its feet but has accumulated too much sagittal velocity.

We also confirmed that the VHIP stabilizer performs similarly its LIP counterpart~\cite{caron2019icra} for nominal walking, with a sleight height increase at the final braking step.

\section{Conclusion}

We proposed a linear feedback controller for the variable-height inverted pendulum based on best-effort pole placement under input feasibility and state viability constraints. This solution is simple to implement, coincides with the 3D DCM for small perturbations and does not require any additional parameter. It can recover from larger perturbations than the 3D DCM by leveraging both the ankle and height-variation strategies.

\section*{Acknowledgments}

The author warmly thanks Arnaud Tanguy for his assistance with experiments, as well as Niels Dehio, Johannes Englsberger, Gabriel Garc\'{i}a and Robert Griffin for their feedback on previous versions of this manuscript.

\addtolength{\textheight}{-1.8cm}

\bibliographystyle{unsrt}
\bibliography{refs}

\end{document}